\documentclass{article}

\usepackage{arxiv}

\usepackage[utf8]{inputenc} 
\usepackage[T1]{fontenc}    
\usepackage{hyperref}       
\usepackage{url}            
\usepackage{booktabs}       
\usepackage{amsfonts}       
\usepackage{nicefrac}       
\usepackage{microtype}      
\usepackage{lipsum}
\usepackage{graphicx}
\graphicspath{ {./images/} }

\usepackage{xurl}
\usepackage{xcolor}

\title{Approaches to Semantic Textual Similarity in Slovak Language: From Algorithms to Transformers}

\author{
 Lukas Radosky\textsuperscript{0000-0003-3909-3219} \\
  Department of Applied Informatics\\
  Faculty of Mathematics, Physics and Informatics\\
  Comenius University Bratislava\\
  Bratislava, Slovakia \\
  \texttt{lukas.radosky@fmph.uniba.sk} \\
   \And
 Miroslav Blstak\textsuperscript{0000-0002-8437-8769} \\
  Kempelen Institute of Intelligent Technologies\\
  Bratislava, Slovakia \\
  \texttt{miroslav.blstak@kinit.sk} \\
   \And
 Matej Krajcovic\textsuperscript{0009-0001-7706-1333} \\
  Department of Applied Informatics\\
  Faculty of Mathematics, Physics and Informatics\\
  Comenius University Bratislava\\
  Bratislava, Slovakia \\
  \texttt{krajcovic70@uniba.sk} \\
   \And
 Ivan Polasek\textsuperscript{0000-0001-6004-701X} \\
  Department of Applied Informatics\\
  Faculty of Mathematics, Physics and Informatics\\
  Comenius University Bratislava\\
  Bratislava, Slovakia \\
  \texttt{ivan.polasek@fmph.uniba.sk} \\
}

\begin{document}
\maketitle

\noindent\colorbox{yellow!20}{\parbox{\linewidth}{
  This is a preprint of a paper that was presented at the \textbf{IEEE 24th World Symposium on Applied Machine Intelligence and Informatics (SAMI 2026)}.
}}

\begin{abstract}
Semantic textual similarity (STS) plays a crucial role in many natural language processing tasks. While extensively studied in high-resource languages, STS remains challenging for under-resourced languages such as Slovak. This paper presents a comparative evaluation of sentence-level STS methods applied to Slovak, including traditional algorithms, supervised machine learning models, and third-party deep learning tools. We trained several machine learning models using outputs from traditional algorithms as features, with feature selection and hyperparameter tuning jointly guided by artificial bee colony optimization. Finally, we evaluated several third-party tools, including fine-tuned model by CloudNLP, OpenAI's embedding models, GPT-4 model, and pretrained SlovakBERT model. Our findings highlight the trade-offs between different approaches.
\end{abstract}


\section{Introduction}
\label{sec:Intro}
Semantic textual similarity (STS) measures how similar or dissimilar two chunks of text are. Higher STS scores indicate a greater degree of semantic similarity between the texts. STS is a key metric in natural language processing (NLP) tasks, including machine translation, summarization, and question answering~\cite{2016_Agirre_SemevalTask1}.

Like many NLP tasks, STS receives relatively little attention in low-resourced languages. This paper focuses on one such language, Slovak. We evaluate selected STS algorithms on a corpus of Slovak texts. The resulting similarity scores are used as features to train several machine learning (ML) models. Hyperparameter tuning and feature selection are performed jointly using the artificial bee colony (ABC) optimization algorithm. We assess the performance of available pretrained AI models, traditional STS algorithms, and our custom ML models.

The main contributions of this paper are:

\begin{itemize}
    \item an overview of available STS algorithms and tools for Slovak texts,
    \item an empirical evaluation of these methods,
    \item practical conclusions on the applicability of these methods, including guidance on preferred use cases.
\end{itemize}

The rest of the paper is organized as follows. Section~\ref{sec:RelatedWork} reviews related work and provides an overview of the current state of research. Section~\ref{sec:EvalUnsupervised} presents our evaluation of traditional STS algorithms. Section~\ref{sec:EvalSupervised} describes the training process and evaluation of the ML models. Third-party tool evaluation is discussed in Section~\ref{sec:EvalThirdParty}. Finally, Section~\ref{sec:Conclusion} summarizes the key findings of this paper.

\section{Related work}
\label{sec:RelatedWork}

Traditional STS algorithms include string-based, statistical (corpus-based), and knowledge-based approaches. In contrast, AI-based approaches often train ML models using the outputs of traditional algorithms combined with other textual features.

String-based algorithms focus on the lexical structure of text without considering its underlying semantics. These are typically divided into character-based and term-based approaches. Character-based algorithms treat text as a sequence of individual characters, disregarding higher-level linguistic structure. Common examples include longest common subsequence~\cite{1974_WagnerFischer_TheString2StringCorrectionProblem} and Levenshtein~\cite{1965_Levenshtein_BinaryCodesCapableOfCorrectingSpuriousInsertionsAndDeletionsOfOnes} distance. Term-based algorithms measure STS at the word level. Some treat the text as a bag of words, e.g. Sorensen-Dice coefficient~\cite{Dice_1945_MeasuresOT} or Jaccard coefficient~\cite{1901_Jaccard}. With the text properly tokenized, these approaches can also be applied at the character level. Other term-based approaches rely on numerical vector representations of text, including Euclidean~\cite{1997_Friedman_OnBiasVariance} and cosine~\cite{1946_Bhattacharyya_OnAMeasureOfDivergence} distance.

Corpus-based algorithms leverage statistical information from large text corpora to capture semantic associations between words. Typically, they construct numerical vector representation of the text and compute STS using vector distance metrics, most often cosine similarity. Notable approaches include Hyperspace Analogue to Language~\cite{1996_LundBurgess_ProducingHighDimensionalSemanticSpace,1995_LundBurgessAtchley_SemanticAndAssociativePriming} and DISCO~\cite{2009_Kolb_DISCO}.

Knowledge-based algorithms use semantic networks such as WordNet~\cite{1995_Miller_Wordnet} to represent word meanings. Edge-based knowledge algorithms utilize graph representations of word semantics in the semantic network. These include Leacock-Chodorow~\cite{1998_LeacockChodorow} and Wu-Palmer~\cite{1994_WuPalmer} algorithms. Some knowledge-based algorithms enhance edge-based information with statistical features of words, i.e. information content. These include Resnik~\cite{1995_Resnik} and Lin~\cite{1998_Lin} algorithms. Overlap-based algorithms focus on word definition overlap, obtaining definitions from a WordNet or a Wikipedia page. These include adapted Lesk~\cite{2002_AdaptedLesk} and Hirst-St. Onge~\cite{1995_HirstStOnge} algorithms.

Outputs from multiple traditional algorithms can be combined for enhanced performance. A common approach is to train a ML model - such as ridge regression~\cite{2015_SemEval_DLS} or support vector regression~\cite{2015_SemEval_ExbThemis} - on the similarity scores produced by these algorithms.

In recent years, the focus of STS-related research has shifted toward semantic networks, ontologies, and word embeddings, sparking interest in corpus-based approaches~\cite{2021_Iqbal_WordEmbeddingBasedTextualSemanticSimilarityMeasureInBengali,2022_Teixeira_ComparisonOfSemanticSimilarityModelsOnConstrainedScenarios,2021_Zhou_SemanticSimilarityOfInverseMorphemeWordsBasedOnWordEmbedding}, knowledge-based approaches~\cite{2022_Almarsoomi_ANewFeatureBasedMethodForSimilarityMeasurementUnderTheLinuxOperatingSystem,2022_Slater_EvaluatingSemanticSimilarityMethodsForComparisonOfTextDerivedPhenotypeProfiles,2022_Sousa_TheSupervisedSemanticSimilarityToolkit} and neural networks~\cite{2022_Ji_AShortTextSimilarityCalculationMethodCombiningSemanticAndHeadWordAttentionMechanism,2022_Sadipour_PersianSentences,2021_Zad_ASurveyOfDeepLearningMethodsOnSemanticSimilarityAndSentenceModeling}.

One dominant modern approach to STS is the use of language models. Although they primarily rely on statistical features, they have become a prevailing paradigm in NLP due to their strong empirical performance and ease of use. However, their limited interpretability and susceptibility to hidden biases~\cite{2021_Bender_OnTheDangersofStochasticParrotsCanLanguageModelsBeTooBig} may hinder deployment in sensitive applications. Despite these concerns, they are widely used in STS-related research~\cite{2022_Patricoski_AnEvaluationOfPretrainedBertModelsForComparingSemanticSimilarityAcrossUnstructuredClinicalTrialTexts,2022_Sun_SentenceSimilarityBasedOnContexts,2022_Yamagiwa_ImprovingWordMoversDistanceByLeveragingSelfAttentionMatrix}.

For Slovak, multilingual models such as XLM-R~\cite{2020_Conneau_XLMR} or WikiBERT~\cite{2021_Pyysalo_WikiBert} are available. Additionally, a Slovak-specific model, SlovakBERT~\cite{2021_pikuliak_slovakbert}, has been developed using a web-crawled Slovak corpus, offering improved performance for Slovak-language STS tasks.

\section{Traditional algorithm evaluation}
\label{sec:EvalUnsupervised}

\begin{figure*}[htbp]
\includegraphics[width=\textwidth]{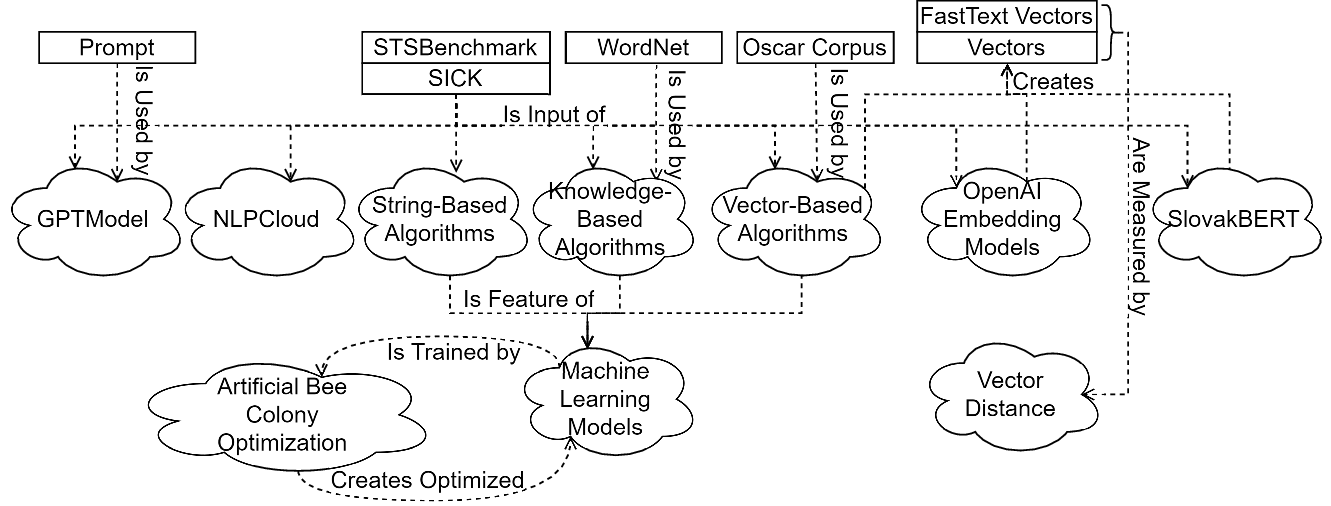}
\caption{Overview of the evaluation pipeline, including preprocessing, optimization, model training, and third-party tool evaluation.} \label{fig_activity}
\end{figure*}

Since there are no datasets specifically designed for STS in Slovak, we employed machine-translated versions of the STS Benchmark~\cite{2017_semeval} and SICK~\cite{2014_sick_a,2014_sick_b} datasets. Translation was performed using the \textit{googletrans}\footnote{\url{https://pypi.org/project/googletrans/}} Python library. While machine translations are not always correct, machine-translated texts are used in many real-world use cases, which is why we opted for this approach. To maintain comparability with existing research, we evaluated STS algorithms separately on each dataset. Evaluation was based on Pearson correlation, following the standard set by SemEval workshops and many other STS-related papers. The complete processing pipeline used for this paper is illustrated in Fig.~\ref{fig_activity}.

Given the morphological richness of Slovak, we also conducted some parallel evaluations on lemmatized versions of the datasets, using the UDPipe lemmatizer~\cite{2016_lemma_udpipe,2017_lemma_udpipe}\footnote{\url{https://github.com/jwijffels/udpipe.models.ud.2.5/blob/master/inst/udpipe-ud-2.5-191206/slovak-snk-ud-2.5-191206.udpipe}}.

Term-based STS algorithms treat text as a set of words, though they can also be adapted to operate on character-level representations. We evaluated both configurations and reported the better-performing variant in Table~\ref{tab:STS_scores}. The performance differences between word-based and character-based modes were not statistically significant. Among these algorithms, the Ochiai coefficient achieved the highest score.

\begin{table}[htbp]
  \centering
  \caption{Pearson correlation scores achieved by traditional algorithms, our machine learning models, and third-party tools on STS Benchmark and SICK datasets. For parametrizable methods, the correlation score of the best-performing configuration is reported. For our ML models and the fine-tuned SlovakBERT, scores reflect performance on validation subsets of the respective datasets.}
  \begin{tabular}{|l|c|c|}
    \hline
    & \textbf{STS Benchmark} &  \textbf{SICK} \\
    \hline
    \hline
    \textbf{Character-based STS algorithms} &&           \\
    \hline
    Hamming~\cite{1950_Hamming_ErrorDetectingAndErrorCorrectingCodes}     & 0.233 & 0.325           \\
    Levenshtein~\cite{1965_Levenshtein_BinaryCodesCapableOfCorrectingSpuriousInsertionsAndDeletionsOfOnes}     & 0.492 & 0.487           \\
    Damerau-Levenshtein~\cite{1964_Demerau_ATechniqueForComputerDetectionAndCorrectionOfSpellingErrors}     & 0.485 & 0.487           \\
    Jaro~\cite{1995_Jaro_ProbabilisticLinkageOfLargePublicHealth,1989_Jaro_AdvancesInRecordLinkageMethodology}     & 0.523 & 0.509          \\
    Jaro-Winkler~\cite{1990_Winkler_StringComparatorMetrics}      & 0.502 & 0.497           \\
    Needleman-Wunsch~\cite{1970_Needleman_AGeneralMethodApplicableToTheSearchForSimilarities}     & 0.397 & 0.452           \\
    Smith-Waterman~\cite{1981_SmithWaterman_IdentificationOfCommonMolecular}     & 0.456 & 0.467           \\
    LCSSeq~\cite{1974_WagnerFischer_TheString2StringCorrectionProblem}     & \textbf{0.532} & \textbf{0.516}           \\
    LCSStr~\cite{2013_Gomaa_ASurveyOfTextSimilarityApproaches}     & 0.491 & 0.491           \\
    \hline
    \textbf{Term-based STS algorithms} &&           \\
    \hline
    Jaccard~\cite{1901_Jaccard}     & 0.568 & 0.527           \\
    Sorensen-Dice~\cite{Dice_1945_MeasuresOT}     & 0.562 & 0.541           \\
    Overlap~\cite{2016_MajumderGoutamPakrayParthaGelbukh_SemanticTextualSimilarityMethodsToolsAndApplicationsASurvey}     & 0.560 & 0.495           \\
    Cosine     & 0.562 & 0.532           \\
    Ochiai~\cite{1936_ochiai}     & \textbf{0.580} & \textbf{0.550}           \\
    \hline
    \textbf{Statistical STS algorithms} &&           \\
    \hline
    HAL~\cite{1996_LundBurgess_ProducingHighDimensionalSemanticSpace}     & 0.126 & 0.220           \\
    ESA~\cite{2007_GabrilovichMarkovitch_ExplicitSemAnalysis}     & 0.319 & 0.343          \\
    DISCO~\cite{2009_Kolb_DISCO}     & 0.347 & 0.366           \\
    FastText~\cite{2018_fast_text}      & 0.420 & 0.498           \\
    OpenAI word-level embeddings     & \textbf{0.547} & \textbf{0.554}           \\
    \hline
    \textbf{Knowledge-based STS algorithms} &&           \\
    \hline
    Wu-Palmer~\cite{1994_WuPalmer}     & 0.184 & 0.237           \\
    Path~\cite{2018_Akila_SemanticSimilarity}     & \textbf{0.265} & \textbf{0.328}          \\
    Leacock-Chodorow~\cite{1998_LeacockChodorow}     & 0.231 & 0.278           \\
    \hline
    \hline
    \textbf{Custom ML models} &&           \\
    \hline
    Linear Regression     & 0.671 & 0.652           \\
    Bayesian Ridge Regression     & 0.672 & 0.650           \\
    Support Vector Regression     & 0.648 & 0.629           \\
    Decision Tree Regression     & 0.635 & 0.643           \\
    Random Forest Regression     & 0.654 & 0.655           \\
    Gradient Boosting Regression     & \textbf{0.685} & \textbf{0.702}           \\
    XGBoost     & 0.678 & 0.696           \\
    \hline
    \hline
    \textbf{Third-party tools} &&           \\
    \hline
    OpenAI full-text embeddings     & 0.756 & 0.718           \\
    GPT-4     & 0.780 & 0.740           \\
    NLPCloud     & \textbf{0.824} & \textbf{0.778}           \\
    SlovakBERT     & 0.7537 & 0.7515           \\
    \hline
    
  \end{tabular}
  \label{tab:STS_scores}
\end{table}

\clearpage

To construct vectors for HAL, ESA, and DISCO, we used the Slovak version of the OSCAR corpus\footnote{\url{https://huggingface.co/datasets/oscar-corpus/OSCAR-2201/tree/main/compressed/sk\_meta}}~\cite{2020_oscarsk_a,2020_oscarsk_b}. We also evaluated existing FastText vectors\footnote{\url{https://huggingface.co/facebook/fasttext-sk-vectors}} and vectors generated by OpenAI embedding models\footnote{\url{https://platform.openai.com/docs/guides/embeddings}}. Since HAL, ESA, DISCO, and FastText produce word-level vectors, we used the OpenAI embedding models to generate word-level vectors as well in this phase. All combinations of vector construction algorithms and similarity metrics were tested. As shown in Table~\ref{tab:STS_scores}, OpenAI embeddings outperformed all the other statistical algorithms, followed by FastText vectors.

Knowledge-based algorithms extract word meaning from a semantic network. We used an existing Slovak WordNet~\cite{2013_slovak_wordnet} obtained from the Open Multilingual WordNet (OMW) collection\footnote{\url{https://github.com/goodmami/wn}}. Several approaches for aggregating word-level similarity scores into sentence-level similarity scores were tested. Pearson correlations of the best-performing strategies are reported in Table~\ref{tab:STS_scores}.

In summary, string-based algorithms, particularly term-based methods, achieved the highest correlation scores. These were followed by statistical approaches, with OpenAI and FastText embeddings achieving the highest scores in that category. Knowledge-based algorithms consistently underperformed, with statistically significant differences compared to the other algorithm families.

\section{Machine learning model evaluation}
\label{sec:EvalSupervised}

The outputs of traditional STS algorithms were used as features for a set of supervised machine learning (ML) models, with the task formulated as a regression problem. The models included linear, Bayesian ridge, support vector, decision tree, random forest, gradient boosting, and XGBoost regression. We used model implementations provided by the \textit{sklearn}~\cite{2011_sklearn} and \textit{xgboost}\footnote{\url{https://xgboost.readthedocs.io}} Python libraries.

Feature selection and hyperparameter tuning were jointly performed using the Artificial Bee Colony (ABC) optimization algorithm, implemented via the \textit{hive}~\cite{2017_abc_hive}\footnote{\url{https://github.com/rwuilbercq/Hive}} library. To maintain experimental reproducibility and manage computational cost, each model–dataset pair was optimized over 30 iterations with a population size of 50.

Each candidate solution (or agent) in the ABC algorithm was represented by a numerical vector consisting of two parts: one for feature selection and the other for hyperparameter tuning. For each traditional STS algorithm, a corresponding value in the feature-selection part determined whether it would be included as a feature (with a 50\% probability threshold). If applicable, the same value also controlled the parameter configuration of the traditional algorithm to be used. The hyperparameter section encoded values for the ML model's parameters, drawn from a predefined search space. A single solution defines a specific model configuration—its selected features and hyperparameter values.

The fitness function was defined as the average Pearson correlation across a stratified 10-fold cross-validation, using 80\% of the dataset. The remaining 20\% was reserved for final validation. A separate optimization run was performed for each combination of ML model (e.g. linear regression), dataset (e.g. SICK), and dataset version (e.g. lemmatized).

After the ABC algorithm identified the optimal feature set and hyperparameter values for each configuration, we retrained the corresponding model on the full training set and evaluated it on the held-out validation set. The final Pearson correlation scores are reported in Table~\ref{tab:STS_scores}.

\section{Third-party tool evaluation}
\label{sec:EvalThirdParty}

In addition to the traditional algorithms and our custom ML models, we also evaluated a range of third-party tools and pretrained models on the STS task. The results of these evaluations are presented in the "Third-party tools" section of Table~\ref{tab:STS_scores}.

We first tested OpenAI’s embedding models—\textit{text-embedding-3-small}, \textit{text-embedding-3-large}, and \textit{text-embedding-ada-002}—by embedding full sentences (as opposed to the previously mentioned word-level embeddings) in each dataset and computing vector similarities using Manhattan, Euclidean, Minkowski ($p=3$), and cosine distances. Across all datasets, the \textit{text-embedding-3-large} model combined with cosine distance consistently yielded the highest correlation scores: 0.756 for STS Benchmark and 0.718 for SICK. However, statistical tests did not confirm a significant performance difference across the models or similarity metrics. In both cases, the Pearson score is higher than for any of the ML models we trained and the word-level OpenAI embeddings. This is likely due to the broader statistical knowledge available to the large models, as well as the richer contextual information captured by full sentences compared to individual words.

We also evaluated the \textit{GPT-4} model. For each sentence pair in the dataset, the model was prompted with the instruction: \textit{What is semantics text similarity of those two sentences? They are in slovak language and please respond in number float or int on scale 0-5}. Several prompt variants were tested on smaller data samples, and the above prompt produced the most consistent results. In occasional cases where the output did not conform to the expected numeric format, the prompt was simply reissued. This approach achieved Pearson correlations of 0.780 on STS Benchmark and 0.740 on SICK—outperforming all of OpenAI's embedding models.

The commercial NLPCloud tool\footnote{\url{https://nlpcloud.com/}} achieved the highest Pearson scores in our evaluation: 0.824 for STS Benchmark and 0.778 for SICK. These results were obtained using its \textit{Paraphrase Multilingual Mpnet Base V2} model fine-tuned for STS, which is based on the sentence-BERT architecture~\cite{2019_bert}\footnote{\url{https://huggingface.co/sentence-transformers/paraphrase-multilingual-mpnet-base-v2}}.

Finally, we evaluated the open-source SlovakBERT model~\cite{2021_pikuliak_slovakbert} by fine-tuning it on a portion of the STS Benchmark dataset and evaluating it on the remaining portion as well as on the SICK dataset. The fine-tuned SlovakBERT achieved Pearson scores of approximately 0.75 on both datasets, performing comparably to the best OpenAI embedding models.

In summary, the NLPCloud tool outperformed all other evaluated third-party systems. \textit{GPT-4} model also produced strong results, exceeding those of the embedding models. The performance of the fine-tuned SlovakBERT model was similar to that of the best OpenAI models, demonstrating the effectiveness of domain-specific fine-tuning.

\section{Conclusion}
\label{sec:Conclusion}

In this article, we focused on solving the problem of semantic text similarity specifically for the Slovak language, as it still does not receive sufficient attention compared to major languages (such as English). We presented a comparative evaluation of traditional semantic textual similarity algorithms, custom machine learning models, and third-party tools on Slovak versions of the STS Benchmark and SICK datasets.

Among traditional methods, term-based algorithms consistently achieved the highest performance. Statistical algorithms performed similarly well when high-quality word vectors were used in combination with cosine similarity. Knowledge-based algorithms underperformed, potentially due to the limitations of the Slovak WordNet. As such, we currently discourage their use for Slovak.

Our supervised ML models, which used outputs from traditional STS algorithms as features, were optimized using the Artificial Bee Colony algorithm with 10-fold cross-validation as the fitness measure. Gradient Boosting Regression and XGBoost emerged as the top-performing models, with XGBoost offering the additional benefit of shorter training time. Lemmatization provided marginal gains, but its overall impact was limited.

Third-party tools leveraging deep learning outperformed both traditional algorithms and our ML models. The NLPCloud tool achieved the highest Pearson scores overall, making it a strong candidate for production use when a paid solution is acceptable. Alternatively, fine-tuning the open-source SlovakBERT model proved to be an effective and cost-free option, yielding performance comparable to state-of-the-art commercial tools.

While third-party tools deliver superior results and are easy to deploy, they often come with financial or computational costs. In contrast, traditional and custom ML approaches are freely available and maintainable after deployment but typically achieve lower correlation scores. Each method presents a different trade-off, and the optimal choice depends on specific application requirements and resource constraints.

For future work, we encourage further adaptation of advanced STS techniques and the development of high-quality linguistic resources (currently, WordNet) for low-resource languages such as Slovak.

\section*{Acknowledgment}

This research and paper was 100\% funded by the EU NextGenerationEU through the Recovery and Resilience Plan for Slovakia under the project "InnovAIte Slovakia, Illuminating Pathways for AI-Driven Breakthroughs" No. 09I02-03-V01-00029.

The authors would like to thank their colleagues and students
at the Comenius University in Bratislava and Kempelen Institute of Intelligent Technologies for their suggestions
and discussion.

\bibliographystyle{unsrt}  
\bibliography{references}

\end{document}